\documentclass{article}

\usepackage{spconf}
\usepackage{graphicx} 
\usepackage{algorithm}
\usepackage{algpseudocode}
\usepackage{LA}

\usepackage{subcaption}
\usepackage{comment}
\usepackage{hyperref}

\usepackage{afterpage}

\title{Understanding Neural Network Systems for Image Analysis using Vector Spaces}
\twoauthors{Rebecca Pattichis}{Dept. of Computer Science \\ University of California, Los Angeles \\ Los Angeles CA, USA \\
pattichi@g.ucla.edu}{Marios S. Pattichis}{
     Dept. of Electrical and Computer Engineering \\
     University of New Mexico \\
     Albuquerque NM, USA \\
     pattichi@unm.edu}

\begin{document}
\maketitle
\begin{abstract}
   There is strong interest in developing mathematical methods
        that can be used to understand complex neural networks
        used in image analysis.
    In this paper, we introduce techniques from Linear Algebra 
        to model neural network layers
        as maps between signal spaces.
    First, we demonstrate how signal spaces can be used to 
       visualize weight spaces and 
       convolutional layer kernels.
    We also demonstrate how
       residual vector spaces can be used to further visualize
       information lost at each layer.
    Second, we study invertible networks using 
        vector spaces for computing input images that yield specific outputs.
    We demonstrate our approach on two 
       invertible networks and ResNet18.\footnote{Code can be accessed at \url{https://github.com/rpattichis/SSIAI_VectorSpaces}.}
\end{abstract}
\section{Introduction}
While neural networks systems 
    perform extremely well in image analysis tasks,
    there is still a lack of understanding of 
    which image representations are best
    captured by different layers.
With their increasing size and integration into 
    important applications (e.g., biomedical \cite{XAI}), 
    it is critical for models to become interpretable. 
The goal of this paper is to 
    suggest methods for understanding neural
    networks based on the use of
    vector spaces and Linear Algebra.

Earlier efforts on visualizing neural network layers suggested
    determining inputs that maximize the activation function
    for a given input image \cite{vis2009}.
In  \cite{saliency2014}, the authors introduced the use
    of saliency maps that allowed us to verify that
    specific regions of an image contributed to its classification score.
In a recent survey \cite{survey},
    the authors summarize many efforts to interpret neural networks,
    including the standard practice of visualizing the convolution filters.

Our focus differs from prior approaches
     by focusing on developing layer interpretations based on the 
     four fundamental vector spaces associated with the weight matrix.
We develop signal and residual (rejected signal) spaces
    to understand how input images get transformed into output images
    and what image components are removed from each layer.
While the concept of invertible neural networks (INNs) have been previously studied (e.g., \cite{vis2009}), we study invertibility using vector spaces.
Similar to \cite{saliency2014}, we compute input vectors
    for different network outputs based on a variety of methods.
We demonstrate our approach on invertible neural networks and
    ResNet18, a more complex network.

The rest of the paper is broken into three sections.
We describe the methodology in section \ref{sec:methods}, 
   provide results in section \ref{sec:results}, and give 
   concluding remarks in \ref{sec:conclusion}.

\section{Methods}\label{sec:methods}
We define the four fundamental spaces in section \ref{sec:four}.
We then proceed to provide interpretation of weight vectors
    using projections in section \ref{sec:proj}.
We extend our approach to weight matrices in section \ref{sec:matrices}.
We then consider computing input image vectors that produce
    desirable outputs in section \ref{sec:inverse}.

\subsection{The four fundamental signal spaces}\label{sec:four}
Let $x$ denote the flattened vector input to a neural network layer.
Here, $x$ can represent an input image (or video).
However, for the purposes of what we are trying to show, we view $x$ as a column vector.
We model the output using:
\begin{equation}
     \text{Out} = f(W x + \text{bias}),
\end{equation}
where
      $W$ denotes the weight matrix,
      bias denotes the bias vector,
      $f$  denotes the activation function, and
      Out denotes the output vector.
The bias term can be absorbed in $W'x'$, 
    where $x'$ extends $x$ by adding 1 in the last element, and $W'$ extends $W$ by adding the bias term.
In what follows, without loss of generality,
    we will ignore the bias term.
Here, we note that convolutions represent special
       cases of the weight matrix.       
Alternatively, for CNN layers, we consider
       signal spaces on the convolution kernels themselves.
      
For the purposes of our development, we define $y$ using $y = W x$.
We make sense of $y = W x$ using
          four fundamental spaces.   
We define the \textbf{signal space} using:
   \SignalDef
The signal space represents what $W$ interprets as the signal component of $x$.
After processing the signal, we are led to study the column space, 
      redefined here as the \textbf{signal output space} 
      $W x$ given by:\\
      \SignalOutDef
The \textbf{signal output space} 
       represents the set of output images that we can reach
       for any given input image.

We now have the elegant equation that maps the signal images
      $x_{\text{signal}} \in \Signal $
      to the reachable output images
      $y_{\text{signal-out}} \in  \SignalOut$:
\begin{equation} 
  W x_{\text{signal}} = y_{\text{signal-out}}. 
     \label{eq:basic}
\end{equation}         
We note that the pseudoinverse $W^+$
      solves equation \eqref{eq:basic}
      exactly using $x_{\text{signal}} = W^+ y_{\text{signal-out}}$.
              
In contrast to the signal space, we define the
     \textbf{rejected signal space} in terms of the null-space of 
     $W$:
    \RejSignalDef
We note that the rejected signal space
    describes all of the input images that have no impact
    on the outputs!
Similarly, in contrast to the output space, we define
     the \textbf{rejected output space} using
     the left null space of $W$:
    \RejSignalOutDef
The input image space is decomposed into 
   the signal and rejected signal spaces  
   as given by  \cite{strang2022}
\begin{align*}
   \mathbf{R}^n &= \mathbf{ColumnSpace}(W^T) \oplus \mathbf{NullSpace}(W)         \\
                         &= \Signal \oplus \RejSignal. 
\end{align*}                          
The output image space is decomposed into the 
       signal output space and the rejected signal output space 
       as given by:                  
\begin{align*}                          
   \mathbf{R}^m &= \mathbf{ColumnSpace}(W)  \oplus  \mathbf{NullSpace}(W^T)      \\
                          &= \SignalOut \oplus \RejSignalOut.  
\end{align*}

\subsection{Understanding weight vectors using projections}\label{sec:proj} 
Let $w$ denote the weight column vector associated with a single neuron.
In this case, $w^T x$ lies in the $\Signal[W=w^T]$ space.
Here, the projection of $x$ onto $w$ is a scaled version of $w^T x$ given by:
  $$ p = \left((w^T x) / \norm{w}^2 \right) x,
     \quad \norm{w}^2 = w^T w. $$
In this case, the weight vector completely removes
  the signal components that belong to $\RejSignal[W=w^T]$ given by:
\begin{equation}
    \text{residual} = x - p \quad\text{satisfying}\quad (x-p)^T w = 0,
    \label{eq:residual}
\end{equation}
where $\text{residual}$ refers to the residual image component
   that is ignored by the weight vector!
In terms of explainability,
    we want to examine
    the $\text{residual}$ image to make sure that
    it does not contain any important signal components.
It is also interesting to note that the input
    image energy is distributed between
    the projection on $\Signal[W=w^T]$ and 
    $\RejSignal[W=w^T]$ spaces as given by:
    $\norm{x}^2 = \norm{p}^2 + \norm{\text{residual}}^2$. 
Thus, we can also measure the amount of image energy
    removed by the residual using:
    $\norm{\text{residual}}^2 / \norm{x}^2$.

\begin{figure*}[t!]
     \centering
     \begin{subfigure}[b]{0.8\textwidth}
         \centering
         \includegraphics[width=\textwidth]{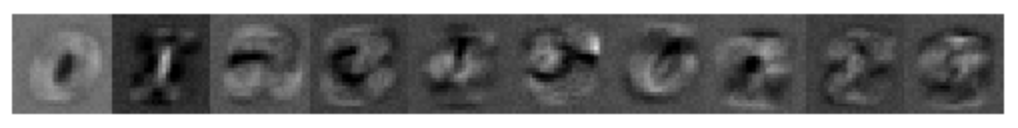}
     \end{subfigure}
     \begin{subfigure}[b]{0.8\textwidth}
         \centering
         \includegraphics[width=\textwidth]{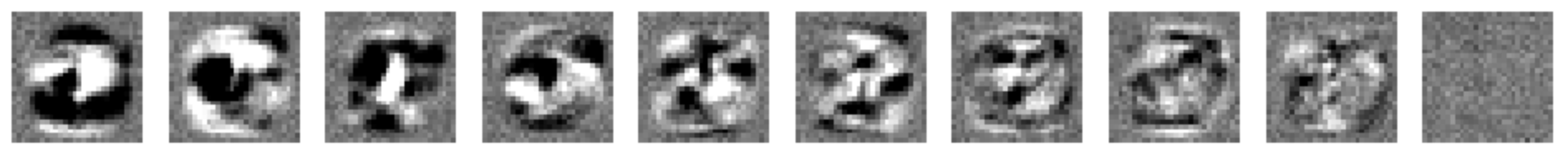}
     \end{subfigure}
     \begin{subfigure}[b]{0.8\textwidth}
         \centering
         \includegraphics[width=\textwidth]{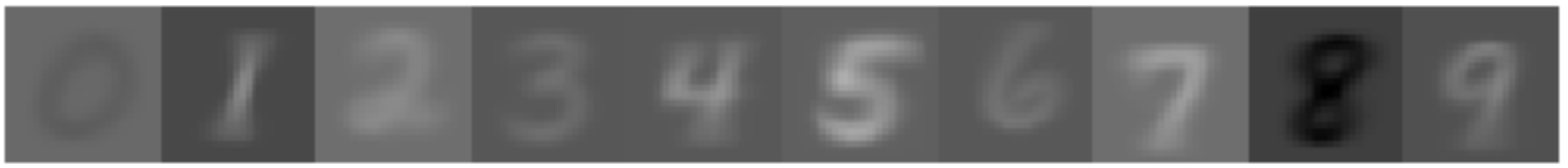}
     \end{subfigure}
     \caption{
     Vector spaces for a single-layer fully connected neural network applied to the 
         MNIST digits dataset.    
     The top row represents the original weights.
     The middle row represents the signal space:
         $\sigma_0 v_0, \dots, \sigma_9 v_9$.
     The condition number is 7.22. 
     The last row represents the residual vectors
          when the network is applied to the average of each digit class
          (see equation (\ref{eq:residual})).}
     \label{fig:single-layer}
\end{figure*}

\subsection{Understanding weight matrices using signal spaces}\label{sec:matrices}
We compute the signal spaces using the Singular Value Decomposition:
  $ W = U \Sigma V^T$.
Along the diagonal of $\Sigma$, we have the associated
  singular values:
   $\sigma_1 \geq \sigma_2 \geq \dots \geq \sigma_r \geq 0,$
where
   $r$ denotes the rank of the matrix and
   $\sigma_1/\sigma_r$ denotes the condition number of $W$.
Here, we note that a low condition number (near 1) is required
   for stable decompositions.
A high condition number would indicate instability in
   the signal space decomposition.
   
The four fundamental spaces are built using
    the eigenvector decompositions of the
    symmetric matrices $W W^T$ and $W^T W$.
We map inputs in $\Signal$ to $\SignalOut$
   using the unit eigenvectors of each space as
   given by:
  \begin{equation}
      \sigma_1 v_1,  
        \dots, \sigma_r v_r,
        \label{eq:signalvis}
  \end{equation}
where we can visualize the relative
   importance of each eigenvector
   through its corresponding singular vector.
More generally, we have:
\begin{align*}
   x &= x_{\Signal} + x_{\RejSignal}, \\
\end{align*}	
where the components are given using:
\begin{align*} 
    x_{\Signal} &= 
        a_1 v_1 + a_2 v_2 + \dots + a_r v_r  \\
    x_{\RejSignal} &= 
        a_{r+1} v_{r+1} + \dots + a_n v_n
\end{align*}         
and \,$a_i = v_i^T x$\, gives the coefficients.

\subsubsection{Simplified interpretation of convolutional layers}
We consider a simplified interpretation of convolutional layers using
   the convolution kernels.
Here, we replace the rows of $W$ with the flattened convolutional kernels.
Hence, our signal spaces refer to the mapping between the support of
    each kernel and the output pixels.

\subsection{Input image generation based on ideal outputs and invertible networks}\label{sec:inverse}
We consider the problem of computing inverse maps given
   desirable outputs.
We note that vector spaces offer a solution
   provided that we use activation functions that are fully invertible
   (e.g., SELU, $\tanh(.)$, sigmoid).
In this case, we can iteratively invert each layer to recover the signal components using: 
\begin{equation}
    x_{\Signal} = W^+ f^{-1}\left(\text{Out}\right).
    \label{eq:inverse}
\end{equation}
Unfortunately, for non-invertible activation functions, we 
   will need to redefine the vector spaces using convex polytopes.   
More generally, we consider a computational approach
   that is generally applicable to any neural network.
We seek to find the input image that 
    generates the minimum distance from an ideal output.
To obtain realistic images, we 
    use the maxima and minima values achieved
    over the training set to define the ideal output vectors.
For example, the ideal output for the first category would
    have the max value over the first output and the minimum value over the rest of them.

We consider two approaches for estimating input images for ideal outputs.
First, we try to minimize the distance to the ideal outputs
    over the training set.
We define the \textit{avg-img} to be the average image over
    all of the training images for the training class that we are interested in.
We define the \textit{min-img} to be the training
    image that minimizes the distance over the set of all training images.
We generalize the \textit{min-img} using the
    \textit{avg-min-img} that is defined to be the average image
    over the training images that are within the lowest 25th percentile
    of distances to the ideal output.
Similar to \cite{saliency2014}, we also train
    the input layer with frozen weights to see if we
    can produce an input image that is even closer to the ideal output.

\begin{figure*}[t!]
    \centering
    \includegraphics[width=0.75\textwidth]{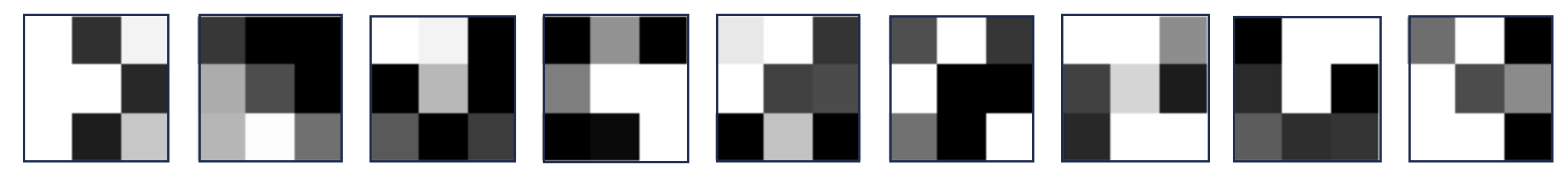}
    \caption{
    The signal space for the first 2D convolution layer in the first Sequential layer of ResNet fine-tuned for MNIST classification (99\% accuracy, 1.07 condition number).}
    \label{fig:resnet-row}
\end{figure*}

\begin{figure*}[t!]
    \centering
    \includegraphics[width=0.9\textwidth]{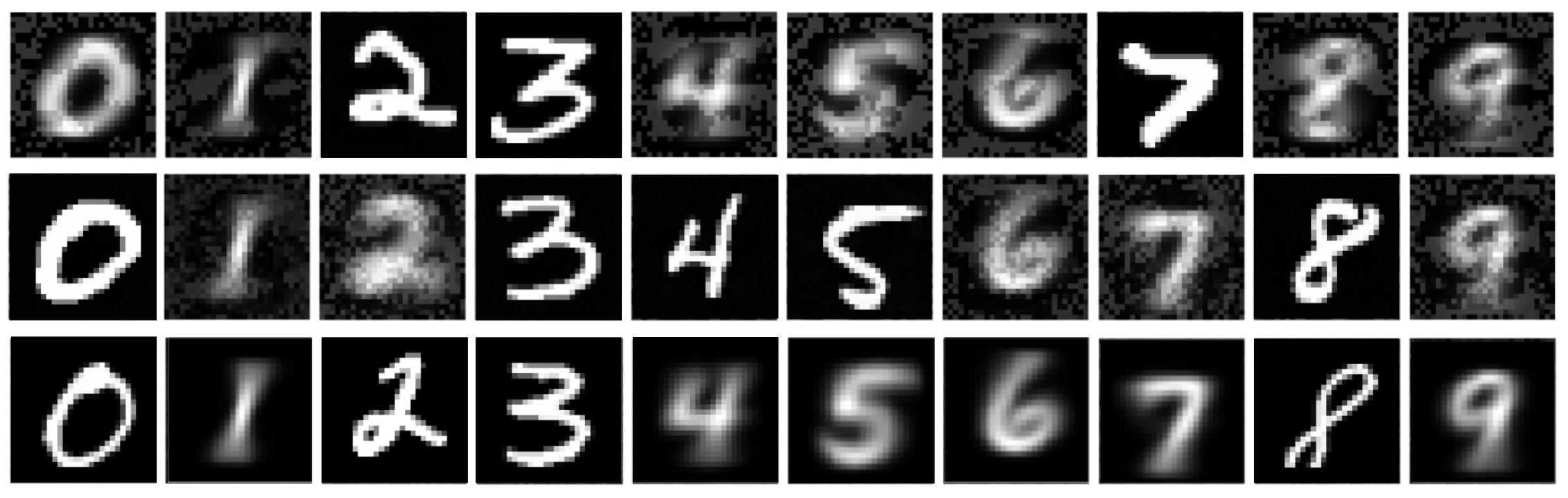}
    \caption{Generated ideal input images for each digit using different algorithms (see section \ref{sec:inverse}).
     Top row: 1-layer FCNN (92\% accuracy): 
        \textit{avg-img}+training for 0, 1, 4, 5, 6, 8, and 9;
        \textit{min-img}+training for 2, 3, and 7.
     Middle row: 5-layer FCNN (97\% accuracy):
        \textit{avg-img}+training for 1, 2, 6, and 7;
        \textit{min-img}+training for 0, 3, 4, 5, 8, and 9.
     Bottom row: ResNet128 (99\% accuracy): 
        \textit{avg-img} for 1, 4, 5, 6, 7, and 9; 
        \textit{min-img} for 0, 2, 3, and 8 that look
        binarized; 
        \textit{avg-min-img} for 4.}
    \label{resnet_layer_invert}
\end{figure*}

\section{Results}\label{sec:results}
We demonstrate our approach on three neural network architectures 
   using the standard MNIST 10-class classification problem.
First, we consider two fully connected neural networks (FCNN):
   (i) a 1-layer FCNN, and
   (ii) a 5-layer FCNN with output dimensions of
        256, 256, 128, 32, and 10.
We used SELU activation functions so that they
   were invertible as discussed in section
   \ref{sec:inverse}.   
Second, we consider the ResNet18 as an example
   of a more complex neural network architecture.
We trained all three networks using a learning rate of 0.001,
   momentum=0.9 and 20 epochs.
In order of neural network complexity, we
   got high classification accuracies at:
   (i) 92\% for 1-layer FCNN,
   (ii) 97\% for 5-layer FCNN, and
   (iii) 99\% for ResNet18.
In what follows, we provide interpretations
    for different layers of the architectures.

We show three vector spaces for the 1-layer FCNN
   in Fig. \ref{fig:single-layer}.
From the signal space, we can clearly see the decreasing importance
   of the signal vectors.
For example, $\sigma_9 v_9$ represents mostly noise,
   and it is far less bright than $\sigma_0 v_0$ ($\sigma_0/\sigma_9=7.22$).
On the other hand, unlike the weight vectors of the top row,
   the rest of the signal vectors (middle row)
   exhibit strong binarized components, dominated by bright and
   dark regions.
The last row of images shows the residual images for the average digit from each class.
Here, it is interesting to note the effectiveness of the weights for 8.
For 8, the residual image shows a dark eight figure,
    suggesting that the information has been removed,
    as expected.
We see a similar pattern for 0.
The rest of the residual images show strong signal components that 
    are likely due to the lack of translational invariance of the
    network when applied to average vectors
    (e.g., see residual for 1).

We present the signal space for the first convolutional layer of 
    ResNet18 in the first Sequential layer in Fig. \ref{fig:resnet-row}.
We note that this layer consists of $64\times64=4096$ $3 \times 3$ kernels
    that we represent with just 9 signal vectors
    $\sigma_1 v_1, \dots, \sigma_9 v_9$.
We note the strong directional selectivity of the signal kernels.
Similar to the binarized patterns of the middle row of Fig. \ref{fig:single-layer},
    we find pixel dominance in different locations or directions.
For example, we have  
    left vertical column dominance in $\sigma_1 v_1$,
    single pixel (lower-center) dominance in $\sigma_2 v_2$,
    top and bottom row dominance in $\sigma_7 v_7$, and
    lower-left diagonal dominance in $\sigma_9 v_9$.
Furthermore, since the condition number is 1.07, it is clear
     that the signal kernels are of equal importance.

We generate input images for each network and each category 
    in Fig. \ref{resnet_layer_invert}.
For ResNet, we note that training did not improve the images
    that were generated by \textit{avg-img}, \textit{min-img}, \textit{avg-min-img}.
This explains why the ResNet images appear either binarized (\textit{min-img})
    or blurry (for \textit{avg-img}, \textit{avg-min-img}).
On the other hand, our low-complexity networks
    proved much easier to train.
Either way, it is clear that our approach of initializing based on the
   original training images proved very effective.
%


\section{Conclusion}\label{sec:conclusion}
The paper introduced the use of the four fundamental vector spaces to understand
    how weight spaces and residual spaces map input images to output images.
Weight spaces represent image (signal) content that gets mapped to the output images.
Residual spaces represent the rejected image content that does not propagate
    through the network.
The paper also discussed invertible networks and 
     methods that estimate input images that yield specific outputs.
In future research, it will be interesting to explore
     if invertible networks can match
     the performance of non-invertible networks.
We note that invertible networks allow us to easily backproject
     output vector spaces to input image spaces.

\section{Acknowledgment}
This work was supported in part by the National Science Foundation under Grant No. 1949230, Nos. 1842220, and 1613637.

\bibliographystyle{IEEEtran}
\bibliography{refs.bib}

\end{document}